\documentclass[letterpaper, 10pt, conference]{ieeeconf}  

\IEEEoverridecommandlockouts                              

\usepackage{graphics} 
\usepackage{epsfig} 
\usepackage{times} 
\usepackage{amsmath} 
\usepackage{amssymb}  
\usepackage[bookmarks=true]{hyperref}
\usepackage{xcolor,colortbl}
\usepackage{color}
\usepackage{siunitx}
\usepackage{multirow}
\usepackage{multicol}
\usepackage[skip=0.3\baselineskip]{caption}
\usepackage{booktabs}
\usepackage{pifont}
\usepackage{cite}
\usepackage{bm}
\usepackage{subcaption}
\usepackage{tikz}
\usetikzlibrary{svg.path}

\usepackage[utf8]{inputenc}
\usepackage{pgfplots}
\DeclareUnicodeCharacter{2212}{−}
\usepgfplotslibrary{groupplots,dateplot}
\usetikzlibrary{patterns,shapes.arrows}
\pgfplotsset{compat=newest}

\newcommand{\bs}{\boldsymbol}

\newcommand{\hide}[1]{}

\DeclareMathOperator*{\argmax}{\arg\!\max}
\definecolor{Gray}{gray}{0.9}
\definecolor{somegray}{rgb}{0.5, 0.5, 0.5}
\newcommand{\darkgrayed}[1]{\textcolor{somegray}{#1}}

\makeatletter
\newcommand*\titleheader[1]{\gdef\@titleheader{#1}}
\AtBeginDocument{%
  \let\st@red@title\@title
  \def\@title{%
    \vskip-3em
    \bgroup\normalfont\large\centering\@titleheader\par\egroup
    \vskip1.5em\st@red@title}
}
\makeatother

\titleheader{ \darkgrayed{This paper has been accepted for publication at the
IEEE/RSJ International Conference on Intelligent \\ Robots and Systems (IROS), Prague, 2021. 
\copyright IEEE} }

\let\oldtwocolumn\twocolumn
\renewcommand\twocolumn[1][]{%
    \oldtwocolumn[{#1}{
    \begin{center}
        \includegraphics[width=0.3\linewidth]{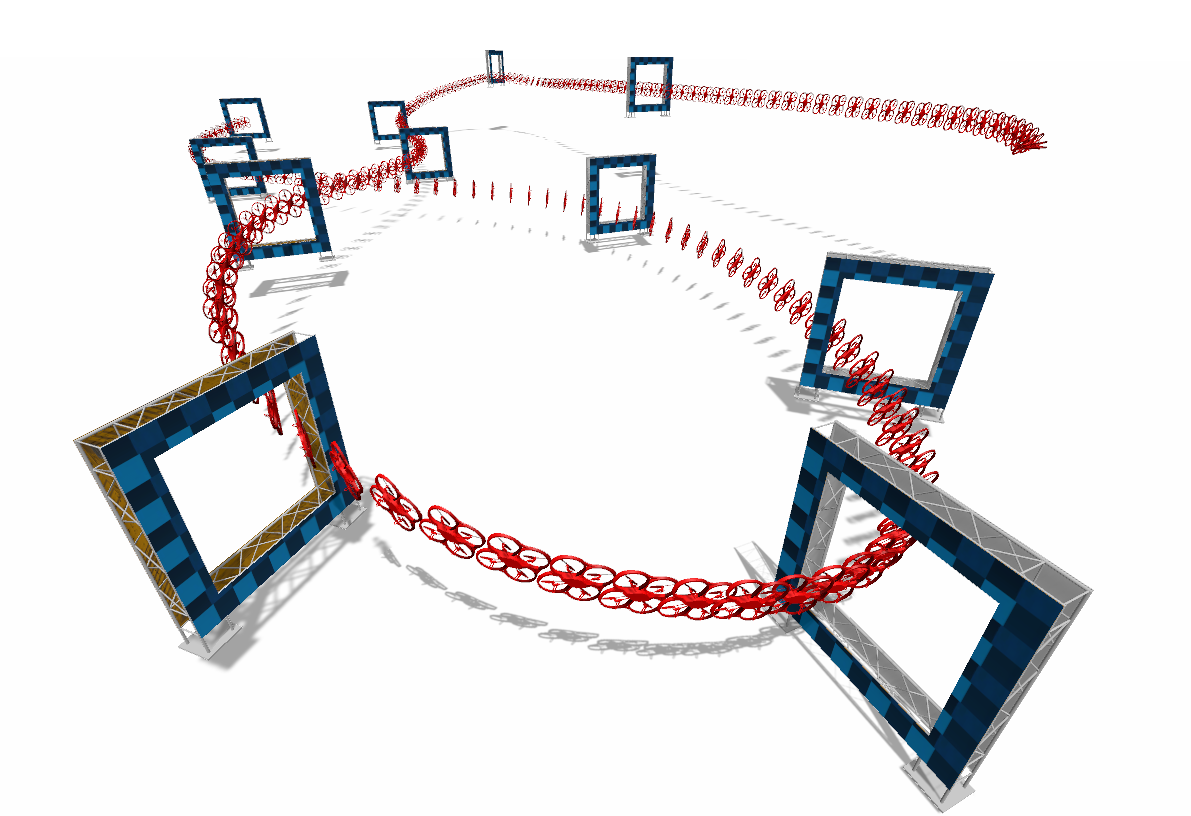}\hfill
        \includegraphics[width=0.32\linewidth]{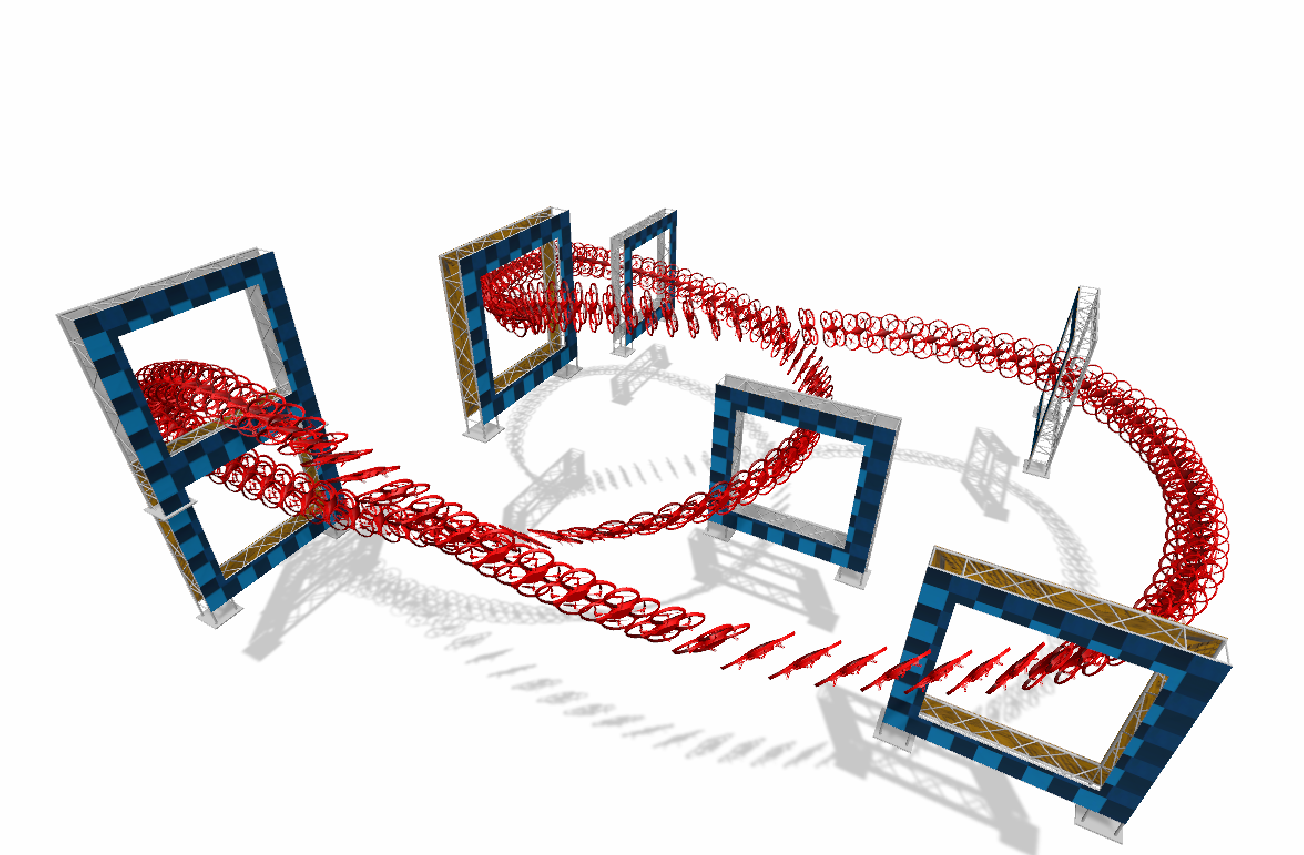}\hfill
        \includegraphics[width=0.33\linewidth]{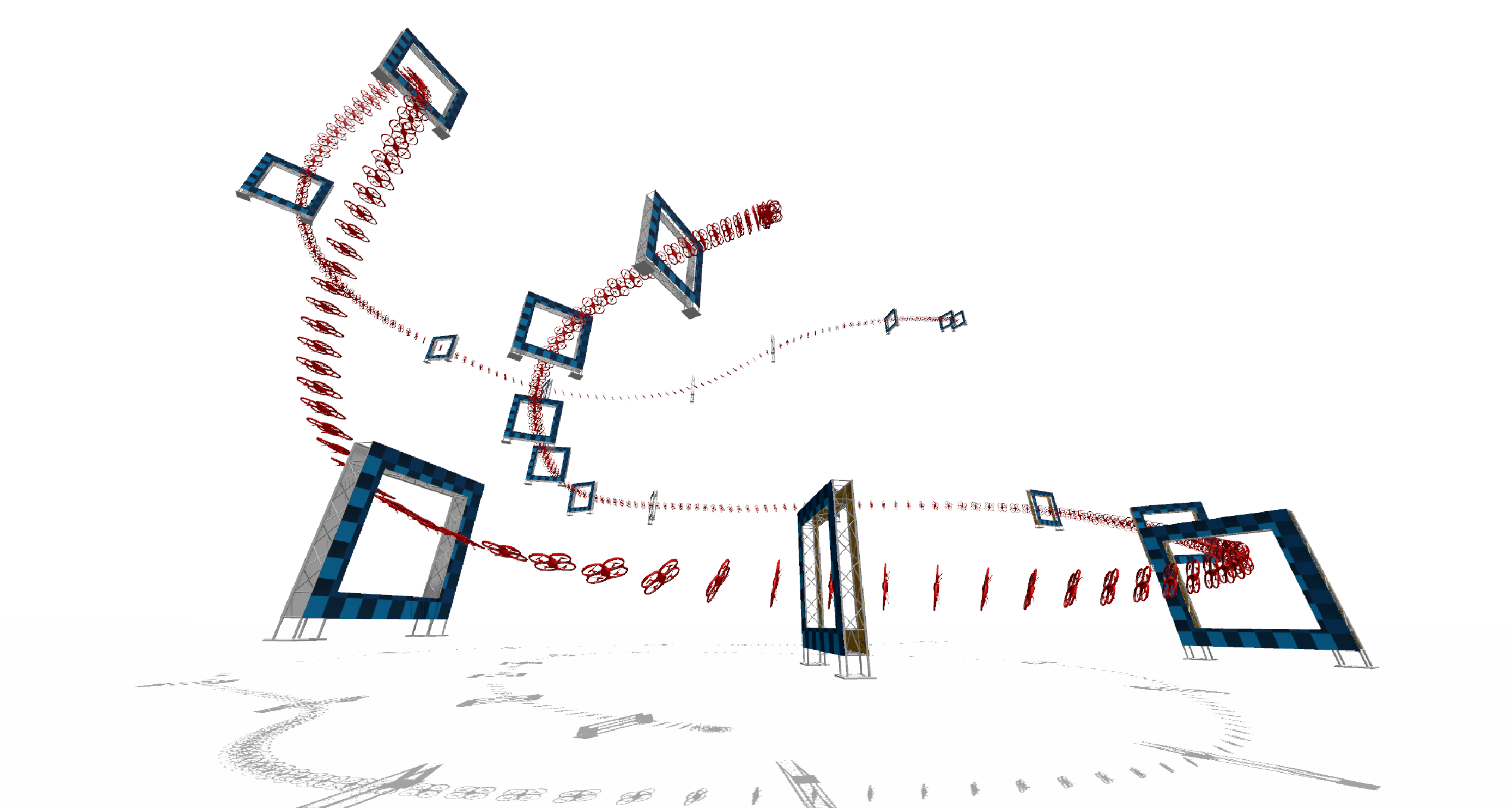}\hfill
        \newline
      \captionof{figure}{A quadrotor navigates at high speed through three different race tracks: \textit{AlphaPilot} (left), \textit{Split-S} (middle), and \textit{AirSim} (right). The trajectories are computed using neural network policies trained with deep reinforcement learning.}
    \end{center}}
    ]
}

\title{\LARGE \bf Autonomous Drone Racing with Deep Reinforcement Learning}

\author{
    Yunlong Song$^{\ast}$,
    Mats Steinweg$^{\ast}$,
    Elia Kaufmann,
    and Davide Scaramuzza 
    \thanks{$^{\ast}$These authors contributed equally.
    The authors are with the Robotics and Perception Group, Department of Informatics, University of Zurich, and Department of Neuroinformatics, University of Zurich and ETH Zurich, Switzerland (\protect\url{http://rpg.ifi.uzh.ch}). M. Steinweg is also with RWTH Aachen University.
    This work was supported by the National Centre of Competence in Research (NCCR) Robotics through the Swiss National Science Foundation (SNSF) and the European Union’s Horizon 2020 Research and Innovation Programme under grant agreement No. 871479 (AERIAL-CORE) and the European Research Council (ERC) under grant agreement No. 864042 (AGILEFLIGHT).
    }
}

\begin{document}

\maketitle
\thispagestyle{empty}
\pagestyle{empty}

\begin{abstract}
In many robotic tasks, such as autonomous drone racing, the goal is to travel through a set of waypoints as fast as possible. 
A key challenge for this task is planning the time-optimal trajectory, which is typically solved by assuming perfect knowledge of the waypoints to pass in advance.
The resulting solution is either highly specialized for a single-track layout, or suboptimal due to simplifying assumptions about the platform dynamics. 
In this work, a new approach to near-time-optimal trajectory generation for quadrotors is presented.
Leveraging deep reinforcement learning and relative gate observations, our approach can compute near-time-optimal trajectories
and adapt the trajectory to environment changes. 
Our method exhibits computational advantages over approaches based on trajectory optimization for non-trivial track configurations. 
The proposed approach is evaluated on a set of race tracks in simulation and the real world, achieving speeds of up to~\SI{60}{\kilo\meter\per\hour} with a physical quadrotor.
\end{abstract}

~\\
\textbf{Video:} \url{https://youtu.be/Hebpmadjqn8}.


\section{Introduction}

In recent years, research on fast navigation of autonomous quadrotors has made tremendous progress, continually pushing the vehicles to more aggressive maneuvers~\cite{foehn2020cpc, loianno2017estimation, kaufmann2020RSS}.
To further advance the field, several competitions have been organized---such as the autonomous drone racing series at the recent IROS and NeurIPS conferences~\cite{moon2019challenges,cocoma2019towards,kaufmann2019beauty,madaan2020airsim} and the AlphaPilot challenge~\cite{guerra2019flightgoggles,foehn2020alphapilot}---with the goal to develop autonomous systems that will eventually outperform expert human pilots. 

Pushing quadrotors to their physical limits raises challenging research problems, including planning a time-optimal trajectory that passes through a sequence of gates. 
Intuitively, such a time-optimal trajectory explores the boundary of the platform's performance envelope.
It is imperative that the planned trajectory satisfies all constraints imposed by the actuators, since the diminishing control authority will otherwise lead to catastrophic crashes.
Furthermore, as quadrotors' linear and angular dynamics are coupled, computing time-optimal trajectories requires a trade-off between maximizing linear and angular accelerations. 

To generate time-optimal quadrotor trajectories, previous work has either relied on numerical optimization by discretizing the trajectory in time~\cite{foehn2020cpc}, expressed the state evolution via polynomials~\cite{ryou2020multi}, or even approximated the quadrotor as a point mass~\cite{foehn2020alphapilot}. 
While the optimization-based method can find trajectories that constantly push the platform to its limits, it requires very long computation times in the order of hours. 
In contrast, approaches that use a polynomial representation~\cite{ryou2020multi} or a point mass approximation~\cite{foehn2020alphapilot} achieve faster computation times, but fail to account for the true actuation limits of the platform, leading to either suboptimal or dynamically infeasible solutions.

In light of its recent successes in the robotics domain~\cite{lee2020learning, hwangbo2017control, gu2017deep}, deep reinforcement learning~(RL) has the potential to solve time-optimal trajectory planning for quadrotors.
Particularly, model-free policy search learns a parametric policy by directly interacting with the environment and is well-suited for complicated problems that are difficult to model.  
Furthermore, policy search allows for optimizing high-capacity neural network policies, which can take different forms of state representations as inputs and allow for flexibility in the algorithm design. 
For example, the learned policy can map raw sensory observations of the robot directly to its motor commands. 
 
Several research questions arise naturally, including how we can tackle the time-optimal trajectory planning problem using deep RL and how the performance compares against model-based approaches.

\subsection*{Contributions}
Our main contribution is a reinforcement-learning-based system that can compute extremely aggressive trajectories that are close to the time-optimal solutions provided by state-of-the-art trajectory optimization~\cite{foehn2020cpc}.
To the best of our knowledge, this is the first learning-based approach for tackling time-optimal quadrotor trajectory planning that is adaptive to environmental changes.
We conduct a large number of experiments, including planning on deterministic tracks, tracks with large uncertainties, and randomly generated tracks.

Our empirical results indicate that: 
1) deep RL can solve the trajectory planning problem effectively, but sacrifices the performance guarantees that trajectory optimization can provide, 
2) the learned neural network policy can handle large track changes and replan trajectories online, which is very challenging and computationally expensive for trajectory-optimization-based approaches,
3) the generalization capability demonstrated by our neural network policy allows for adaptation to different racing environments, suggesting that our policy could be deployed for online time-optimal trajectory generation in dynamic and unstructured environments.
 
The key ingredients of our approach are three-fold: 
1) we use appropriate state representations, namely, the relative gate observations, as the policy inputs, 
2) leveraging a highly parallelized sampling scheme, 
and 3) maximizing a three-dimensional path progress reward, which is a proxy of minimizing the lap time. 

\section{Related Work}
Existing works on time-optimal trajectory planning can be divided into 
sampling-based and optimization-based approaches.
Sampling-based algorithms, such as RRT$^\star$~\cite{karaman2011sampling}, are provably optimal in the limit of infinite samples.
%
As such algorithms rely on the construction of a graph, they are commonly combined with approaches that provide a closed-form solution for state-to-state trajectories. 
Many challenges exist in this line of work; for example, the closed-form solution between two full quadrotor states is difficult to derive analytically when considering the actuator constraints.
For this reason, prior works~\cite{webb2013kinodynamic, allen2019real, foehn2020alphapilot, zhou2019robust} have favoured point-mass approximations or polynomial representations for the state evolution,
which possibly lead to infeasible solutions for the former and inherently sub-optimal solutions for the latter as shown in~\cite{foehn2020cpc}. 

The time-optimal planning for quadrotor waypoints~\cite{foehn2020cpc},
solve the problem by using discrete-time state space representations for the trajectory, and enforcing the system dynamics and thrust limits as constraints. 
A major advantage of optimization-based methods, such as~\cite{spedicato2017minimum, foehn2020cpc, ryou2020multi}, is that they can incorporate nonlinear dynamics and handle a variety of state and input constraints.
For passing through multiple waypoints, the allocation of the traversal times is a priori unknown, rendering the problem formulation significantly more complicated.
The requirement of solving a complex constrained optimization problem
makes most trajectory optimization only suitable for applications where the computation of time-optimal trajectories can be done offline.
%


Deep RL has been applied to autonomous navigation tasks, 
such as high-speed racing~\cite{fuchs2020super, jaritz2018end} and motion planning of autonomous vehicles~\cite{aradi2020survey, roy2002motion}.
The main focus of these methods was to solve navigation problems for wheeled robots, whose motion is constrained to a two-dimensional plane. 
However, finding a time-optimal trajectory for quadrotors is more challenging since the search space is significantly larger due to the high-dimensional state and input space.
Previous works applying RL to quadrotor control have shown successful waypoints following~\cite{hwangbo2017control} and autonomous landing~\cite{rodriguez2019deep}, but neither of these works managed to push the platform to its physical limits. 
%
Our work tackles the problem of racing a quadrotor through a set of waypoints in minimum time using deep RL. 
Our policy does not impose strong priors on the track layout, manifesting its applicability to dynamic environments. 
\section{Methodology}

\subsection{Quadrotor Dynamics}

We model the quadrotor as a 6 degree-of-freedom rigid body of mass $m$ and diagonal moment of inertia matrix $\bm{J}$.
The dynamics of the system can be written as:
\begin{align*}
\mathbf{\dot{p}}_{WB}&=\mathbf{v}_{WB}&\mathbf{\dot{q}}_{WB} &= \frac{1}{2} \mathbf{\Lambda} ( \boldsymbol{\omega}_{B}) \cdot \mathbf{q}_{WB} \\
\mathbf{\dot{v}}_{WB}&=\mathbf{q}_{WB}\odot\mathbf{c}+\mathbf{g} & \boldsymbol{ \dot{\omega}_{B}} &= \mathbf{J}^{-1}(\boldsymbol{\eta} - \boldsymbol{\omega}_{B} \times \mathbf{J} \boldsymbol{\omega}_{B})  
\end{align*}
where $\mathbf{p}_{WB}=[p_x, p_y, p_z]^T$ and $\mathbf{v}_{WB}=[v_x,v_y,v_z]^{T}$ are the position 
and the velocity vectors of the quadrotor in the world frame~${W}$.
We use a unit quaternion $\mathbf{q}_{WB}=[q_{w},q_{x},q_{y},q_{z}]^{T}$ to represent the orientation of the quadrotor
and $\bs{\omega}_{B}= [\omega_x, \omega_y, \omega_z]^T$ to denote the body rates in the body frame~${B}$.
Here, $\mathbf{g}=[0, 0, -g_z]^{T}$ with $g_z=\SI{9.81}{\meter\per\second\squared}$ is the gravity vector,
$\mathbf{J}$ is the inertia matrix, $\boldsymbol{\eta}$ is the three dimensional torque, 
and $\mathbf{\Lambda} (\bs{\omega}_{B})$ is a skew-symmetric matrix.
Finally, $\mathbf{c}=[0, 0, c]^T$ is the mass-normalized thrust vector. 
The conversion of single rotor thrusts $[f_1, f_2, f_3, f_4]$ to mass-normalized thrust $c$ and body torques $\boldsymbol{\eta}$ is formulated as
\begin{align*}
    c = \frac{1}{m} \sum_{i=1}^4 f_i \; ,\quad\boldsymbol{\eta} &= \begin{bmatrix}
    l/\sqrt{2}(f_1 - f_2 - f_3 + f_4)  \\ 
    l/\sqrt{2}(-f_1 - f_2 + f_3 + f_4) \\
    \kappa (f_1 - f_2 + f_3 - f_4) \end{bmatrix} \; ,
\end{align*}
where $m$ is the quadrotor's mass, $\kappa$ is the rotor's torque constant, and $l$ is the arm length.
The full state of the quadrotor is defined as
$\mathbf{x}=[\mathbf{p}_{WB}, \mathbf{q}_{WB}, \mathbf{v}_{WB}, \bs{\omega_{B}}]$ and the control inputs are $\mathbf{u}= [f_1, f_2, f_3, f_4]$. 
We use a 4th order Runge-Kutta method for the numerical integration of the 
dynamic equation, denoted as $f_\text{RK4}(\mathbf{x}, \mathbf{u}, d_t)$, with $d_t$ as the integration time step.  

\subsection{Task Formulation}
We formulate the time-optimal trajectory planning problem in the 
reinforcement learning framework. 
To this end, we model the task using an infinite-horizon Markov Decision Process~(MDP) defined by a tuple $(\mathcal{S}, \mathcal{A}, \mathcal{P}, r, \rho_0, \gamma)$~\cite{sutton1998reinforcement}.
The RL agent starts in a state~$\bs{s}_t \in \mathcal{S}$
drawn from an initial state distribution $\rho_0(\bs{s})$.
At every time step $t$, an action $\bs{a}_t \in \mathcal{A}$ sampled
from a stochastic policy $\pi(\bs{a}_t|\bs{s}_t)$ is executed
and the agent transitions to the next state~$\bs{s}_{t+1}$
with the state transition probability~$\mathcal{P}^{\bs{a}_t}_{\bs{s}_t\bs{s}_{t+1}}
	= \text{Pr}( \bs{s}_{t+1} | \bs{s}_t, \bs{a}_t)$,
receiving an immediate reward~${r(t)\in\mathbb{R}}$. 
Explicit representations of state and action are discussed in Section~\ref{subsec: obs_act}.

The goal of deep RL is to optimize the parameters~$\bm{\theta}$ of a neural network policy~$\pi_{\bm{\theta}}$
such that the trained policy maximizes the expected discounted reward over an infinite horizon. The discrete time-formulation of the objective is 
\begin{equation}
 \pi_{\bm{\theta}}^{\ast} = \argmax_{\pi} \mathbb{E}_{\bm{\tau} \sim \pi} \left[ \sum_{t=0}^{\infty} \gamma^{t} r(t) \right]
\end{equation}
where $\gamma \in [0, 1)$ is the discount factor that trades off long-term against short-term rewards. 
The optimal trajectory~$\bm{\tau}^{\ast}$ is obtained by rolling out the trained policy. 

\subsubsection{Reward Function}
%
%
%
Intuitively, the total time spent on the track directly captures the main objective of the task. 
However, this signal can only be computed upon the successful completion of a full lap, introducing sparsity that considerably increases the difficulty of the credit assignment to individual actions.
A popular approach to circumvent this problem is to use a proxy reward that closely approximates the true performance objective while providing feedback to the agent at every time step. 
In car racing, reward shaping techniques based on the projected progress along the center-line of the race track have shown to work well as an approximation of the true racing reward~\cite{liniger2015optimization, fuchs2020super}.

We extend the concept of a projection-based path progress reward to the task of drone racing by using straight line segments connecting adjacent gate centers as a representation of the track center-line. 
Thus, no additional effort is required for computing the reference path. A race track is completely defined by a set of gates placed in three-dimensional space.

At any point in time, the quadrotor can be associated with a specific line segment based on the next gate to be passed. 
To compute the path progress, we project the position of the quadrotor onto the current line segment. 
Given the current position $\mathbf{p}_\text{c}(t)$ and previous position $\mathbf{p}_\text{c}(t-1)$ of the quadrotor,
we define the progress reward $r_\text{p}(t)$ as follows:
\begin{equation}
    r_\text{p}(t) = s(\mathbf{p}_\text{c} (t)) - s(\mathbf{p}_\text{c} (t-1)).
\end{equation}
Here, $s(\mathbf{p}) = (\mathbf{p} - \mathbf{g}_1) \cdot (\mathbf{g}_2 - \mathbf{g}_1) / \| (\mathbf{g}_2 - \mathbf{g}_1)\|$
defines the progress along the path segment that connects the previous gate's center $\mathbf{g}_1$
with the next gate's center $\mathbf{g}_2$.

\begin{figure}[t]
    \centering
    \resizebox{0.7\linewidth}{!}{
    \input{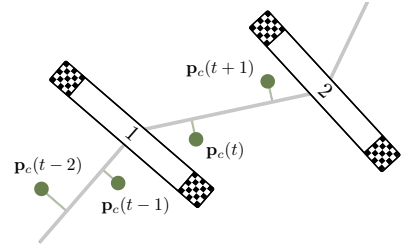}}
    \caption{Illustration of the progress reward. 
    }
    \label{fig:progress_reward}
\end{figure}

\begin{figure}[t]
\centering
\resizebox{0.8\linewidth}{!}{
    \input{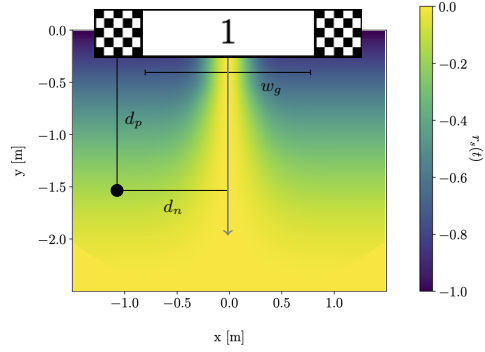}
}
\caption{Illustration of the safety reward. 
}
\label{fig:gate_rewards}
\end{figure}

In addition, we define a safety reward to incentivize the vehicle to fly through the gate center by penalizing small safety margins.
It should be noted that the safety reward is an optional reward component designed to reduce the risk of crashing in training settings that feature large track changes.
The safety reward is defined as follows:
\begin{equation}
    r_\text{s}(t) = - f^2 \cdot \left(1 - \exp \left(- \frac{0.5 \cdot d_n^2}{v} \right)\right)
\end{equation}
where $f = \max \; [1 - (d_p/d_\text{max}), \; 0.0]$ and $v =\max \; [(1 - f) \cdot (w_g/6), \; 0.05]$.  
Here, $d_p$ and $d_n$ denote the distance of the quadrotor to the gate normal and the distance to the gate plane, respectively. 
The distance to the gate normal is normalized by the side length of the rectangular gate $w_g$, while $d_\text{max}$ specifies a threshold on the distance to the gate center in order to activate the safety reward.
We illustrate the safety reward in Figure~\ref{fig:gate_rewards}.

The final reward at each time step $t$ is defined as 
\begin{equation}
r(t) = r_\text{p}(t) + a\cdot r_\text{s}(t) - b\cdot\|\bm{\omega}_t\|^{2} + 
\left\{\begin{matrix}
r_T&\quad \text{if gate collision} \\ 
0  &\text{else} 
\end{matrix}\right.
\end{equation}
where $r_T$ is a terminal reward for penalizing gate collisions 
\begin{equation}
    r_T = - \min \left[ \left(\frac{d_g}{w_g} \right)^{2}, \; 20.0 \right]
\end{equation}
with $d_g$ and $w_g$ denoting the euclidean distance from the position of the crash to the gate center and the side length of the rectangular gate, respectively.
Furthermore, a quadratic penalty on the body rates $\bm{\omega}$ weighted by $b \in \mathbb{R}$ is added at an initial training stage.
Here, $a \in \mathbb{R}$ is a hyperparameter that trades off between progress maximization and risk minimization. 

%

%

\subsubsection{Observation and Action Spaces}
\label{subsec: obs_act}
%
The observation space consists of two main components, one related to the quadrotor's
state~$\mathbf{s}_t^\text{quad}$ and one capturing information about the race track~$\mathbf{s}_t^\text{track}$. 
We define the observation vector for the quadrotor state as ${\mathbf{s}_{t}^\text{quad} = [\mathbf{v}_{WB, t},  \mathbf{\dot{v}}_{WB, t}, \mathbf{R}_{WB, t}, \bm{\omega}_{B, t}] \in \mathbb{R}^{18}}$, which corresponds to the quadrotor's linear velocity, linear acceleration, rotation matrix, and angular velocity, respectively. 
To avoid ambiguities and discontinuities in the representation of the orientation, we use rotation matrices to describe the quadrotor attitude as was done in previous work on quadrotor control using RL~\cite{hwangbo2017control}.
%


We define the track observation vector as $\mathbf{s}_{t}^\text{track}=[\mathbf{o}_1, \alpha_1, \cdots, \mathbf{o}_i , \alpha_i, \cdots], \; i \in [1, \cdots, N]$,
where $\mathbf{o}_i  \in \mathbb{R}^3 $ denotes the observation for gate $i$ 
and $N\in \mathbb{Z}^+$ is the total number of future gates. 
We use spherical coordinates $\mathbf{o}_i = (p_{r}, p_{\theta}, p_{\phi})_i$ to represent the observation of the next gate, as this representation provides a clear separation between the distance to the gate and the flight direction.
For the next gate to be passed, we use a body-centered reference frame, while all other gate observations are recursively expressed in the frame of the previous gate.
Here, $\alpha_i \in \mathbb{R}$ specifies the angle between the gate normal and the vector pointing from the quadrotor to the center of gate $i$. 
The generic formulation of the track observation allows us to incorporate a variable number of future gates into the observation. 
%
%
We apply input normalization using z-scoring by computing observation statistics across deterministic policy rollouts on 
1000 randomly generated three-dimensional race tracks.

The agent is trained to directly map the observation to individual thrust commands, defining the action as $\mathbf{a}_t = [f_1, f_2, f_3, f_4]$. 
Using individual rotor commands allows for maximum versatility and aggressive flight maneuvers.
We use the Tanh activation function at the last layer of the policy network to keep the control commands within a fixed range. 

\begin{figure}[t]
    \centering
    \includegraphics[width=0.9\linewidth]{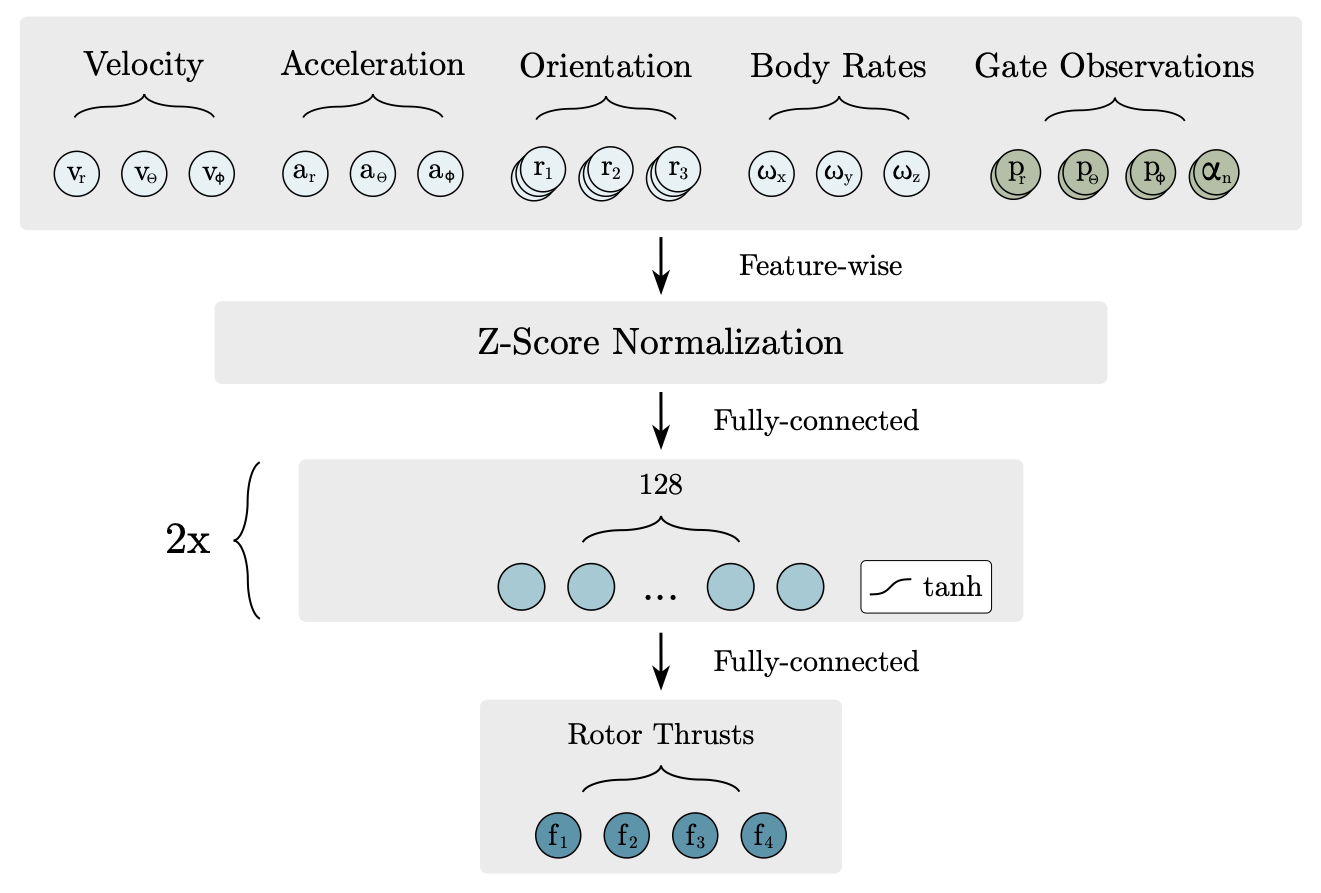}
    \caption{Illustration of the observation vector and network architecture used in this work.}
    \label{fig:catch-eye}
\end{figure}

\subsection{Policy Training}
We train our agent using the Proximal Policy Optimization (PPO) algorithm~\cite{schulman2017proximal}, a first-order policy gradient method that is particularly popular for its good benchmark performances and simplicity in implementation. 
Our task presents challenges for PPO due to the high dimensionality of the search space and the complexity of the required maneuvers.
%
We highlight three key ingredients that allow us to achieve stable and fast training performance on complex three-dimensional race tracks with arbitrary track lengths and numbers of gates:


\begin{figure*}[h!]
\centering
  \includegraphics[width=1.0\textwidth]{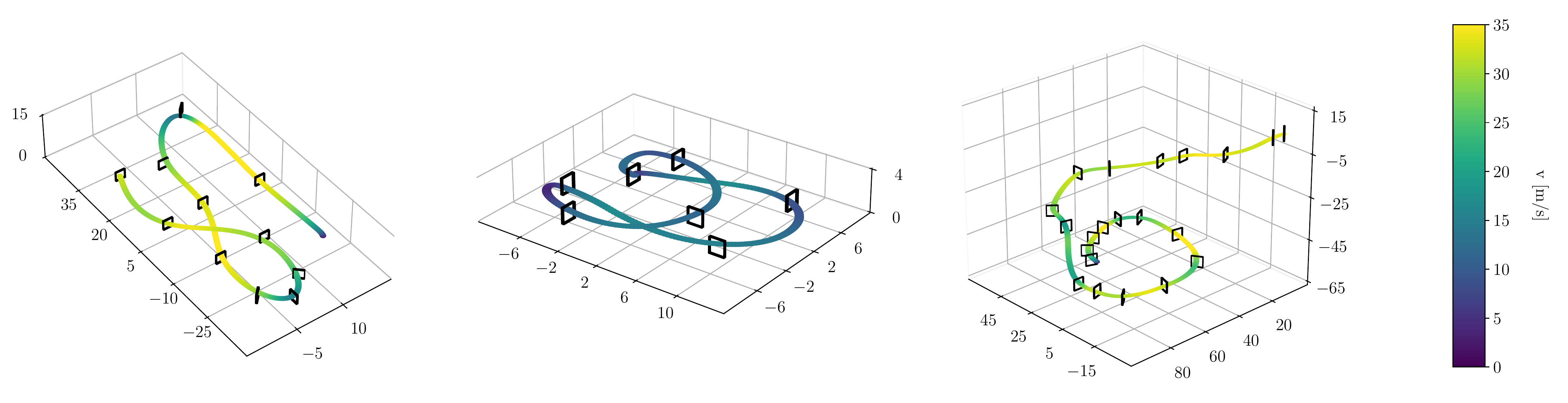}
\caption{Race tracks used for the baseline comparison along with trajectories generated by our approach. 
    From left to right: \textit{AlphaPilot} track, \textit{Split-S} track, \textit{AirSim} track.
    The colors correspond to the quadrotor's velocity.}
\label{fig:race_tracks}
\end{figure*}

\subsubsection{Parallel Sampling Scheme}
Training our agent in simulation allows for performing policy rollouts on up to 100 environments in parallel, resulting in a significant speed-up of the data collection process.
Furthermore, the parallelization can be leveraged to increase the diversity of the collected environment interactions. 
When training on a long race track, initialization strategies can be designed such that rollout trajectories cover the whole race track instead of being limited to certain parts of the track.
Similarly, in a setting where strong track randomization is applied, rollouts can be performed on a variety of tracks using different environments, efficiently avoiding overfitting to a certain track configuration.

\subsubsection{Distributed Initialization Strategy}
%
%
At the early stages of the learning process, a majority of the rollouts terminate on a gate crash.
Consequently, if each quadrotor is initialized at the starting position, the data collection is restricted to a small part of the state space, requiring a large number of update steps until the policy is capable of exploring the full track. 
To counteract this situation, we employ a distributed initialization strategy that ensures a uniform exploration of relevant areas of the state space. 
We randomly initialize the quadrotors in a hover state around the centers of all path segments, immediately exposing them to all gate observations. 
Once the policy has learned to pass gates reliably, we sample initial states from previous trajectories. 
Thus, we retain the benefit of initializing across the whole track while avoiding the negative impact of starting from a hover position in areas of the track that are associated with high speeds. 

\subsubsection{Random Track Curriculum}\label{sec:track_curriculum}
Training a racing policy for a single track already constitutes a challenging problem.
However, given the static nature of the track, an appropriate initialization strategy allows for solving the problem without additional guidance of the training process.
When going beyond the setting in which a deterministic race track is used for training, the complexity of the task increases drastically.
Essentially, training a policy by randomly sampling race tracks, confronts the agent with a new task in each episode.

For this setting, we first design a track generator by concatenating a list of randomly generated gate primitives. 
We mathematically define the track generator as 
\begin{equation}
    \mathbf{T} = [\mathbf{G}_1, \cdots, \mathbf{G}_{j+1}], \quad j \in [1, +\infty)
\end{equation}
where $\mathbf{G}_{j+1}= f(\mathbf{G}_j,   \Delta \mathbf{p}, \Delta \mathbf{R})$ is the gate primitive,
which is parameterized via the relative position $\Delta \mathbf{p}$ and orientation $\Delta \mathbf{R}$ 
between two consecutive gates. 
%
%
Thus, by adjusting the range of relative poses and the total number of gates, we can generate race tracks of arbitrary complexity and length. 
%

To design a training process in which the focus remains on minimizing the lap time despite randomly generated tracks, we propose to automatically adapt the complexity of the sampled tracks based on the agent's performance.
Specifically, we start the training by sampling tracks with only small deviations from a straight line.
The policy quickly learns to pass gates reliably, reducing the percentage of rollouts that results in a gate crash.
We use the crash ratio computed across all parallel environments to adapt the complexity of the sampled tracks by allowing for more diverse relative gate poses.
Thus, the data collection is constrained to a certain class of tracks until a specified performance threshold is reached.
The adaptation of the track complexity can be interpreted as an automatic curriculum, resulting in a learning process that is tailored towards the capabilities of the agent.

\section{Experiments}
We design our experiments to answer the following research questions:
(i) How does the lap time achieved by our learning-based approach compare to optimization algorithms for a deterministic track layout?
(ii) How well can our approach handle changes in the track layout?
(iii) Is it possible to train a policy that can successfully race on a completely unknown track?
(iv) Can the trajectory generated by our policy be executed with a physical quadrotor?

\subsection{Experimental Setup}\label{sec:experimental_setup}
%
%

We use the Flightmare simulator~\cite{song2020flightmare} and implement a vectorized OpenAI gym-style racing environment, which allows for simulating hundreds of quadrotors and race tracks in parallel.
Such a parallelized implementation enables us to collect up to 25000 environment interactions per second during training.
Our PPO implementation is based on~\cite{stable-baselines3}. 
%
%

To benchmark the performance of our method, we use three different race tracks, including a track from the 2019 AlphaPilot challenge~\cite{guerra2019flightgoggles}, a track from the 2019 NeurIPS AirSim Game of Drones challenge~\cite{madaan2020airsim}, and a track designed for real-world experiments (\textit{Split-S}). 
These tracks~(Fig.~\ref{fig:race_tracks}) pose a diverse set of challenges for an autonomous racing drone:
the \textit{AlphaPilot} track features long straight segments and hairpin curves, the \textit{AirSim} track requires performing rapid and very large elevation changes, and the \textit{Split-S} track introduces two vertically stacked gates. 
%
%

To test the robustness of our approach, we use random displacements of the position and the yaw angle of all gates on the \textit{AlphaPilot} track.
Furthermore, we train and evaluate our approach on completely random tracks using the track generator introduced in Section~\ref{sec:track_curriculum}, 
allowing us to study the scalability and generalizability of our approach. 
Finally, we validate our design choices with ablation studies
and execute the generated trajectory on a physical platform. 
All experiments are conducted using a quadrotor configuration with diagonal inertia matrix $\bm{J} = [0.003, 0.003, 0.005] \, \SI{}{\kg\meter\squared}$, torque constant $\kappa = 0.01$ and arm length $\SI{0.17}{m}$.
For the \textit{AlphaPilot} and \textit{AirSim} track, we use a thrust-to-weight ratio of 6.4 while for the \textit{Split-S} track a thrust-to-weight ratio of 3.3 is used (for the real-world deployment).
Unless noted otherwise, we perform all experiments using two future gates in the observation.
We present an ablation study on the number of gates in Section~\ref{sec:ablation}.




\subsection{Baseline Comparison on Deterministic Tracks \label{sec:deterministic-experiment}}
%

We first benchmark the performance of our policy on three deterministic tracks (Section~\ref{sec:experimental_setup}) against two trajectory planning algorithms:
polynomial minimum-snap trajectory generation~\cite{mellinger2011minimum} and optimization-based time-optimal trajectory generation based on complementary progress constraints (CPC)~\cite{foehn2020cpc}. 
Polynomial trajectory generation, with the specific instantiation of minimum-snap trajectory generation, is widely used for quadrotor flight due to its implicit smoothness property.
Although in general a very useful feature, the smoothness property prohibits the exploitation of the full actuator potential at every point in time, rendering the approach sub-optimal for racing.
%
In contrast, the optimization-based method (CPC) provides an asymptotic lower bound on the possibly achievable time for a given race track and vehicle model. 
We do not compare to sampling-based methods due to their simplifying assumptions on the quadrotor dynamics.
To the best of our knowledge, there exist no sampling-based methods that take into account the full state space of the quadrotor.

The results are summarized in Table~\ref{tab:evaluation-policies}.
Our approach achieves a performance within \SI{5.2}{\percent} of the theoretical limit illustrated by CPC.
The baseline computed using polynomial trajectory generation is significantly slower.
Fig.~\ref{fig:race_tracks} illustrates the three evaluated race tracks and the corresponding trajectories 
generated by our approach. 
%
\begin{table}[h]
\caption{Lap time in seconds.}
\label{tab:evaluation-policies}
\begin{center}
    \begin{small}
    \setlength{\tabcolsep}{3pt}
        \begin{tabular}{c||c|c|c}
            \toprule
            \rowcolor{Gray}
            Methods &  AlphaPilot~(\SI{}{\second}) &Split-S~(\SI{}{\second})& AirSim~(\SI{}{\second})   \\
            \hline
            Polynomial~\cite{mellinger2011minimum}  & 12.23& 15.13 & 23.82\\ 
            \hline
            Optimization~\cite{foehn2020cpc} & 8.06 & 6.18 &  11.40 \\
            \hline
            Ours  & 8.14 &  6.50& 11.82 \\
            \bottomrule
        \end{tabular}
    \end{small}
\end{center}
\end{table}

\subsection{Handling Race Track Changes \label{sec:uncertainty-experiment}}
While the results on the deterministic track layouts allow to compare the racing performance to the theoretical limit, such lap times are impossible to obtain in a more realistic setting where the track layout is not exactly known in advance. 

\begin{figure}[t]
\centering
\begin{subfigure}{0.5\linewidth}
\centering
\includegraphics[width=1.0\linewidth,trim=10 0 10 10, clip]{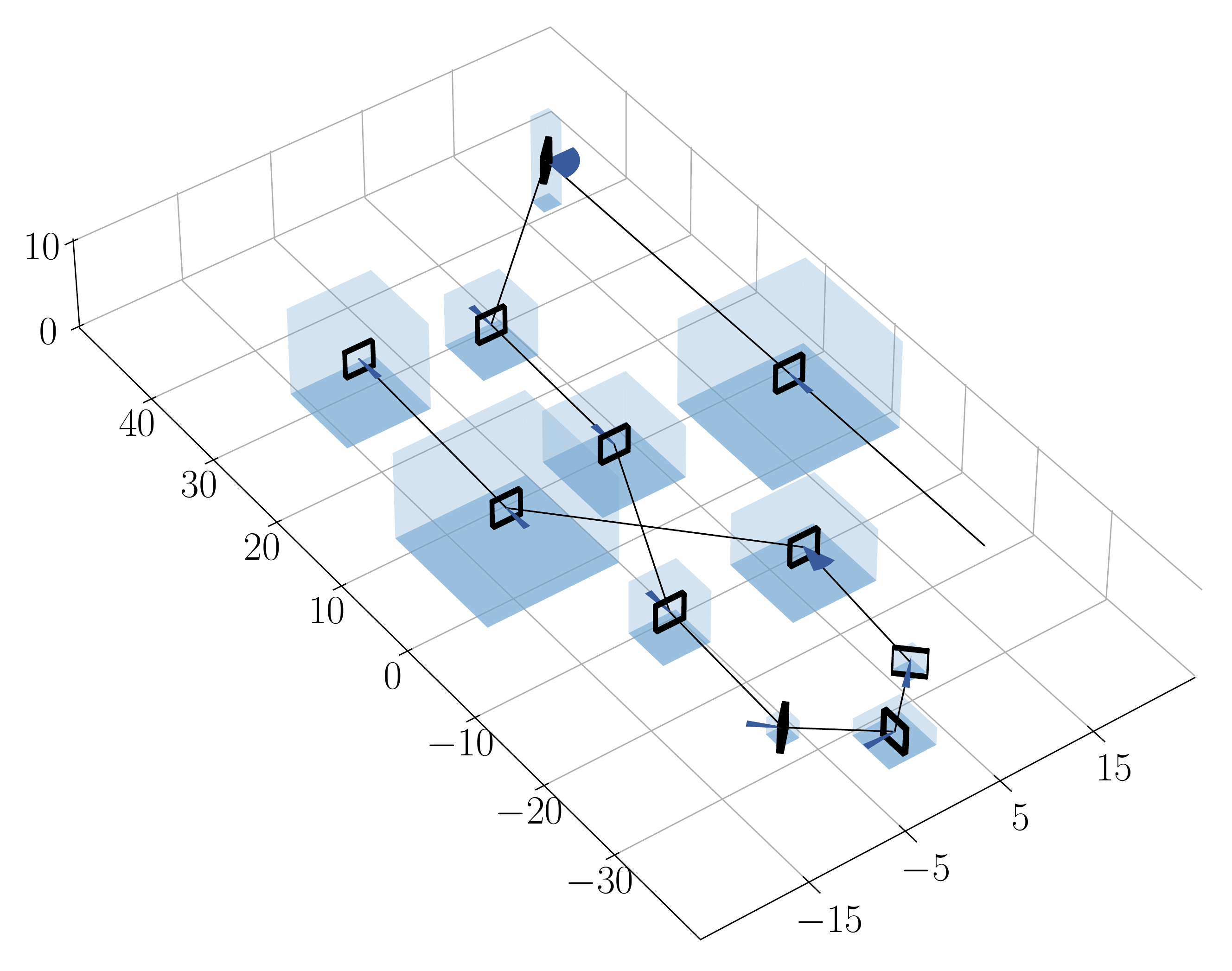}
\end{subfigure}%
\begin{subfigure}{0.5\linewidth}
\centering
\includegraphics[width=1.0\linewidth]{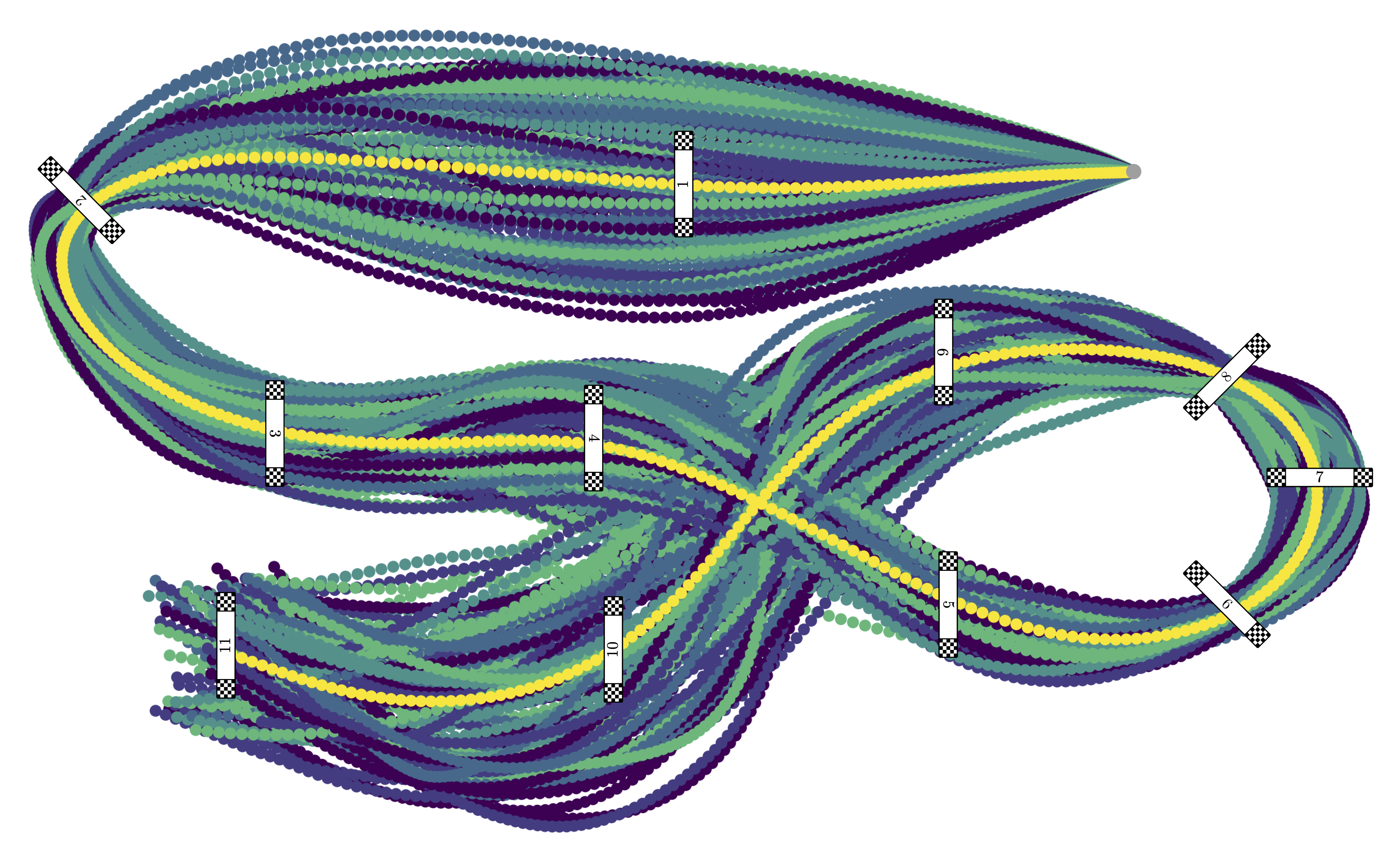}
\end{subfigure}
\caption{\textit{Left:} Track changes for the \textit{AlphaPilot} track. The gate center can be moved to arbitrary positions within the respective cube while possible orientations are illustrated by the blue cones. \textit{Right:} Trajectories on different configurations of the \textit{AlphaPilot} track. The color scheme is chosen to highlight different trajectories and does not reflect any metric related to the trajectory. The nominal trajectory is displayed in yellow.}
\label{fig:perturbed_trajectories_2d}
\end{figure}

We design an experiment to study the robustness of our approach to partially unknown track layouts by specifying random gate displacements on the \textit{AlphaPilot} track from the previous experiment. 
%
Specifically, we use bounded displacements ${[\Delta{x}, \Delta{y}, \Delta{z}, \Delta{\alpha}]}$ on the position of each gate's center as well as the gate's orientation in the horizontal plane.
%
%
Fig.~\ref{fig:perturbed_trajectories_2d} illustrates the significant range of displacements that are applied to the original track layout. 


%
We train a policy from scratch by sampling different gate layouts from a uniform distribution of the individual displacement bounds. 
In total, we create 100 parallel racing environments with different track configurations for each training iteration.  
%
The final model is selected based on the fastest lap time on the nominal version of the \textit{AlphaPilot} track.
We define two metrics to evaluate the performance of the trained policy: the average lap time and the crash ratio.
%
We compute both metrics on three difficulty levels (defined by the randomness of the track), each comprising a set of 1000 unseen randomly-generated track configurations. 
%
%

The results are shown in Table~\ref{tab:displacement-policies}.
The trained policy achieves a success rate of \SI{97.5}{\percent} on the most difficult test set which features maximum gate displacements.
%
%
The lap time performance on the nominal track decreased by \SI{2.21}{\percent} compared to the policy trained without displacements~(Table~\ref{tab:evaluation-policies}). 
\begin{table}[h]
\caption{Evaluation results of handling track uncertainties. The gate randomization level corresponds to the percentage of the maximum gate displacements.}
\label{tab:displacement-policies}
\begin{center}
    \begin{small}
    \setlength{\tabcolsep}{3pt}
        \begin{tabular}{c||c|c}
            \toprule
            \rowcolor{Gray}
           Gate Randomization Level& Avg. Time  (\SI{}{\second}) & \#Crashes   \\
            \hline
            0.00 & 8.32  & 0   \\
            \hline
            0.50 & 8.43  & 0   \\
            \cline{1-3}
            0.75 & 8.55 & 2     \\
            \cline{1-3}
            1.00 & 8.73  & 25   \\
            \bottomrule
        \end{tabular}
    \end{small}
\end{center}
\end{table}
To illustrate the complexity of the task, Fig.~\ref{fig:perturbed_trajectories_2d} shows 100 successful trajectories on randomly sampled tracks using maximum gate displacements along with the nominal trajectory.
%
%

\subsection{Towards Learning a Universal Racing Policy}
We study the scalability and generalizability of our approach by designing an experiment to train a policy that allows us to generate time-optimal trajectories for completely unseen tracks. 

\begin{figure}[h!]
    \centering
    \includegraphics[width=1.0\linewidth]{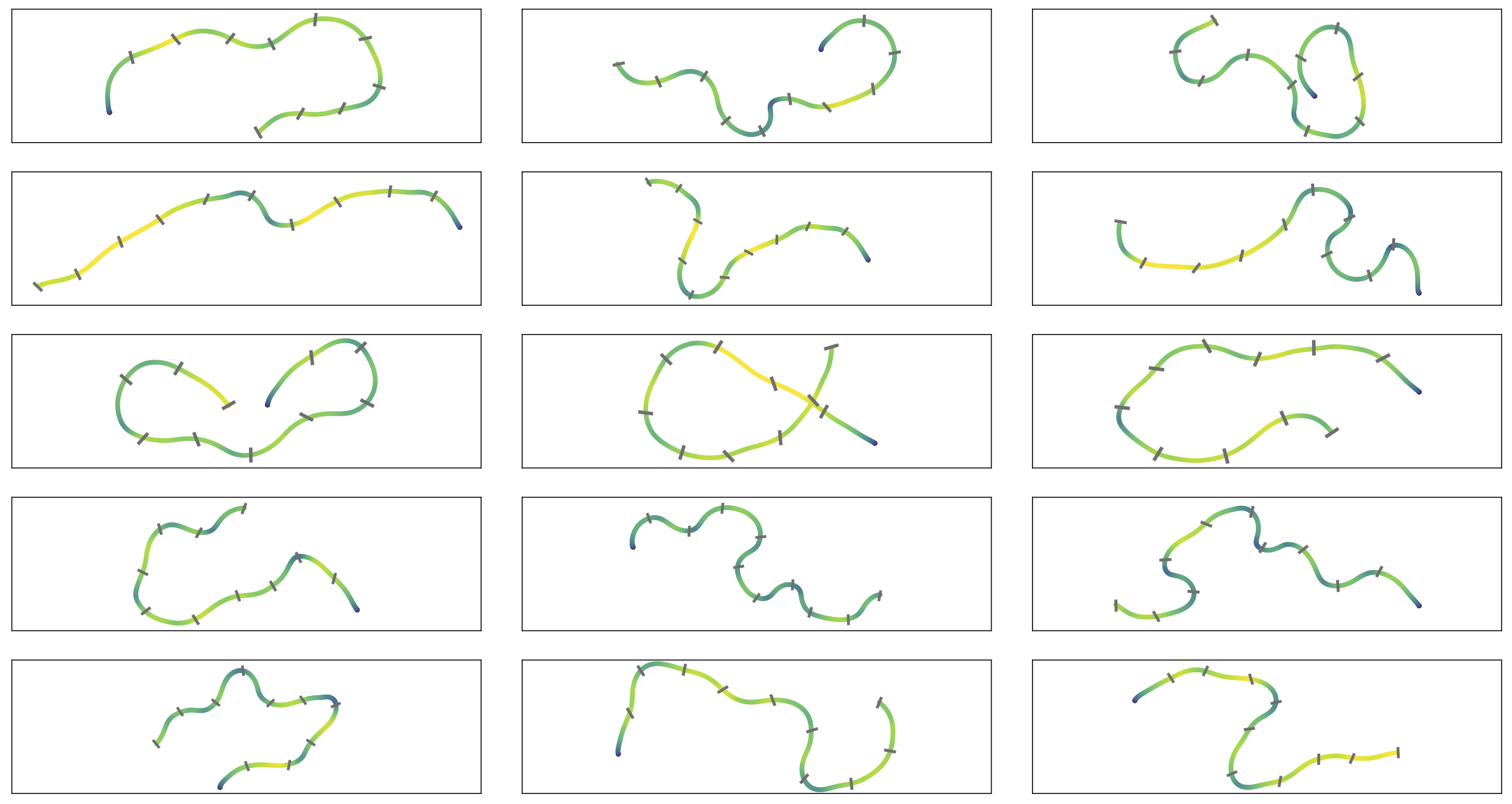}
    \vspace{0.1cm}
    \caption{
    Policy rollouts on randomly generated tracks (visualization only in $xy-$plane). 
    The trajectories range from \SI{110}{m} to \SI{150}{\meter} length with elevation changes of up to \SI{17}{\meter}. 
    }
    \label{fig:random-track-3d}
\end{figure}

To this end, we train three different policies on completely random track layouts in each rollout.
The three policies differ in the training strategy and the reward used: we ablate the usage of the automatic track adaptation and the safety reward.  
%
During training, we continuously evaluate the performance of the current policy on a validation set of 100 unseen randomly sampled tracks.
For each configuration we choose the policy with the lowest crash ratio on the validation set as our final model.
The final evaluation is performed on a test set of 1000 randomly sampled tracks of full complexity and we assess the performance based on the number of gate crashes.

\newcommand{\cmark}{\ding{51}}%
\newcommand{\xmark}{\ding{55}}%

\begin{table}[h!]
\caption{Performance on randomly generated race tracks. We report the number of crashes on 1000 test tracks.}
\label{tab:universal-policies}
\begin{center}
    \begin{small}
    \setlength{\tabcolsep}{3pt}
        \begin{tabular}{c|c|c}
            \toprule
            \rowcolor{Gray}
             Safety Reward & Track Adaptation & \#Crashes  \\
            \hline
            {\color{black} \xmark}  & {\color{black} \xmark} & 71 \\
            \hline
            {\color{black} \xmark}  & {\color{black} \cmark}  & 47  \\
            \hline
            {\color{black} \cmark} & {\color{black} \cmark} & 26 \\
            \bottomrule
        \end{tabular}
    \end{small}
\end{center}
\end{table}
The results of the experiment are displayed in Table~\ref{tab:universal-policies}.
When training the policy without both, the safety reward and the automatic track adaptation, the policy performs worst in terms of number of crashes. 
Adding automatic track adaptation considerably improves performance by reducing the task complexity at the beginning of the training process and taking into account the current learning progress when sampling the random tracks.
%
%
By adding the safety reward, we manage to further reduce the crash ratio, illustrating the positive effect of maximizing the safety margins when training on random tracks.
Our policy achieves \SI{97.4}{\percent} success rate on 1000 randomly generated tracks,
exhibiting strong generalizability and scalability.  

To illustrate the large variety of the randomly sampled tracks, Fig.~\ref{fig:random-track-3d} shows a subset of the evaluation trajectories.

\subsection{Computation Time Comparison and Ablation Studies}

\subsubsection{Computation Time Comparison}
We compare the training time of our reinforcement learning approach to the computation time of the trajectory optimization method in a controlled experiment.
Specifically, we compare the computation times for the three trajectories presented in Section~\ref{sec:deterministic-experiment}, which consists of 7, 11, and 21 gates (
for SplitS, AlphaPilot, and AirSim respectively). 
Also, we evaluate the computation time on randomly generated race tracks consisting of 5, 15, 30 and 35 gates.
For each number of gates we run 3 different experiments and compute the mean computation time. 
We only conduct a single experiment for 35 gates due to the long computation time of the optimization-based approach.
Consequently, we report the computation time for a total of 13 challenging race tracks.

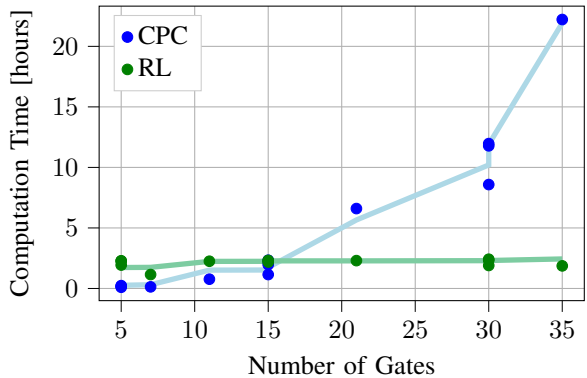
\begin{figure}[t]
    \centering
\begin{tikzpicture}

\definecolor{color0}{rgb}{0.67843137254902,0.847058823529412,0.901960784313726}
\definecolor{color1}{rgb}{ 0.49, 0.8, 0.63}
\definecolor{color3}{rgb}{0.13, 0.6, 0.33}
\definecolor{color4}{rgb}{0.13, 0.38, 0.55}

\begin{axis}[
legend cell align={left},
height=5.5cm,
width=0.93\linewidth,
xmajorgrids,ymajorgrids,
legend style={
  draw opacity=1,
  text opacity=1,
  at={(0.03,0.97)},
  anchor=north west,
  draw=white!80!black
},
tick align=outside,
tick pos=left,
x grid style={white!69.0196078431373!black},
xlabel={Number of Gates},
xmin=3.5, xmax=36.5,
xtick style={color=black},
y grid style={white!69.0196078431373!black},
ylabel={Computation Time [hours]},
ymin=-1.00583333333333, ymax=23.3225,
ytick style={color=black},
ytick={0, 5, 10, 15, 20}
]
\addplot [line width=2pt, color0, forget plot]
table {%
5 -0.0189792512394495
5 0.00763069780053094
5 0.260204041610997
7 0.310412633276859
11 1.5153812804616
15 1.52537573867878
15 1.56198143493899
15 1.73963979444244
21 5.6585400276156
30 10.2011196130975
30 11.4303434694463
30 11.9041744245777
35 21.920842761959
};
\addplot [line width=2pt, color1, forget plot]
table {%
5 1.66185225606159
5 1.66820954752649
5 1.73071329644336
7 1.7436676505663
11 2.23804510171521
15 2.24268759487572
15 2.25551100440344
15 2.27976828567311
21 2.27973667074067
30 2.29608163701793
30 2.30550454215923
30 2.30958550565848
35 2.43863690715845
};
\addplot [draw=blue, fill=blue, mark=*, only marks]
table{%
x  y
5 0.183333333333333
5 0.233333333333333
5 0.1
15 2.33333333333333
15 1.15
15 1.96666666666667
30 11.7833333333333
30 11.9666666666667
30 8.58333333333333
35 22.2166666666667
11 0.766666666666667
21 6.6
7 0.133333333333333
};
\addlegendentry{CPC}
\addplot [draw=green!50!black, fill=green!50!black, mark=*, only marks]
table{%
x  y
5 1.93333333333333
5 2.26666666666667
5 2.28333333333333
15 2.18333333333333
15 2.31666666666667
15 2.25
30 2.33333333333333
30 2.41666666666667
30 1.9
35 1.86666666666667
11 2.25
21 2.3
7 1.15
};
\addlegendentry{RL}
\end{axis}

\end{tikzpicture}
    \caption{Computation time comparison between our learning-based method and the optimization-based method (CPC).  }
    \label{fig:computation-time}
\end{figure}

The results of this experiment are displayed in Figure~\ref{fig:computation-time}.
The results show that the computation time of the optimization-based method increases drastically with the number of gates as the number of decision variables in the optimization problem scales accordingly.
The training time of our learning-based method instead does not depend on the number of gates, making the approach particularly appealing for large-scale environments.  
%

\subsubsection{Ablation Studies\label{sec:ablation}} 

We perform an ablation study to validate design choices of the proposed approach.
Specifically, we focus on the effect of different observation models as well as the safety reward.

\textit{Number of Observed Gates:} Since the number of future gates included in the observation corresponds to neither a fixed distance nor time, we consider it necessary to ablate the effect of the number of gates on the policy's performance.
%
%

%
We train policies on the \textit{AlphaPilot} track using 1, 2 and 3 future gates in the observation.
Both versions of the track, the deterministic layout as well as the layout subject to bounded displacements presented in Section~\ref{sec:uncertainty-experiment} are used.
The final model for evaluation is selected based on the fastest lap time achieved on the nominal track.
The results of this experiment are displayed in Table~\ref{tab:gate-ablation}.
\newcolumntype{C}[1]{>{\centering\arraybackslash}m{#1}}
\begin{table}[h!]
\caption{Results of the gate observation ablation experiment. We report the lap time on the nominal track as well as average lap time
and crash ratio on 1000 test tracks.}
\label{tab:gate-ablation}
\begin{center}
    \begin{small}
    \setlength{\tabcolsep}{5pt}
        \begin{tabular}{C{1.cm}|c|c}
            \toprule
            \rowcolor{Gray}
            \#Gates & Lap Time & Avg. Lap Time (s) /  Crash Ratio (\%)  \\
            \hline
            1 & 8.20 & 9.92 /  23.0 \\
            \hline
            2 & 8.14 & 8.32 / 2.5 \\
            \hline
            3 &  8.16 & 8.36 / 2.3 \\
            \bottomrule
        \end{tabular}
    \end{small}
\end{center}
\end{table}

We show that the final lap time on the deterministic track is not affected significantly by the number of observed gates.
This can be explained by the static nature of the task for which it is possible to overfit to the track layout even without information beyond the next gate to be passed.
However, when introducing gate displacements, using only information about a single future gate is insufficient to achieve fast lap times and minimize the crash ratio.

\textit{Safety Reward:}
We introduce the safety reward in order to alter the performance objective such that the distance to the gate center at the time of the gate crossing is minimized during training.
This becomes relevant in training settings with significant track randomization as well as real-world experiments that require safety margins due to uncertain effects of the physical system. 
%
%
%
We design our experiments to evaluate the impact of two safety reward configurations on the number of crashes as well as the safety margins of successful gate crossings.

\begin{table}[h]
\caption{Results of the safety reward ablation experiment. We report the lap time on the nominal track layout, the crash ratio on 1000 test tracks and statistics about the safety margins of successful gate crossings.}
\label{tab:safety-ablation}
\begin{center}
    \begin{small}
    \setlength{\tabcolsep}{3pt}
        \begin{tabular}{c||c|c|c}
            \toprule
            \rowcolor{Gray}
            $d_{max}$ & Lap Time (s) & Crash Ratio (\%) & Safety Margins (m)  \\
                        \hline
            0.0 &  8.32 & 2.5 & 0.67 $\pm$ 0.28 \\
                        \hline
            2.5 &  8.42 & 0.8 & 0.95 $\pm$ 0.34 \\
                        \hline
            5.0 & 8.48 & 0.6 & 0.94 $\pm$ 0.35 \\
            \bottomrule
        \end{tabular}
    \end{small}
\end{center}
\end{table}
The results of this experiment are displayed in Table~\ref{tab:safety-ablation}.
We show that the safety reward increases the average safety margin for all configurations.
Furthermore, the results indicate that the safety reward consistently reduces the number of crashes on the test set.
However, both configurations of the safety reward result in reduced performance on the nominal track.

\subsection{Real-world Flight}
Finally, we execute the trajectory generated by our policy on
a physical quadrotor to verify the transferability of our policy.
We use a quadrotor that is developed from off-the-shelf components made for drone racing, 
including a carbon-fiber frame, brushless DC motors, 5-inch propellers, 
and a BetaFlight controller.
The quadrotor has a thrust-to-weight ratio of around 4.
\begin{figure}[t]
 \centering
  \includegraphics[width=1.0\linewidth]{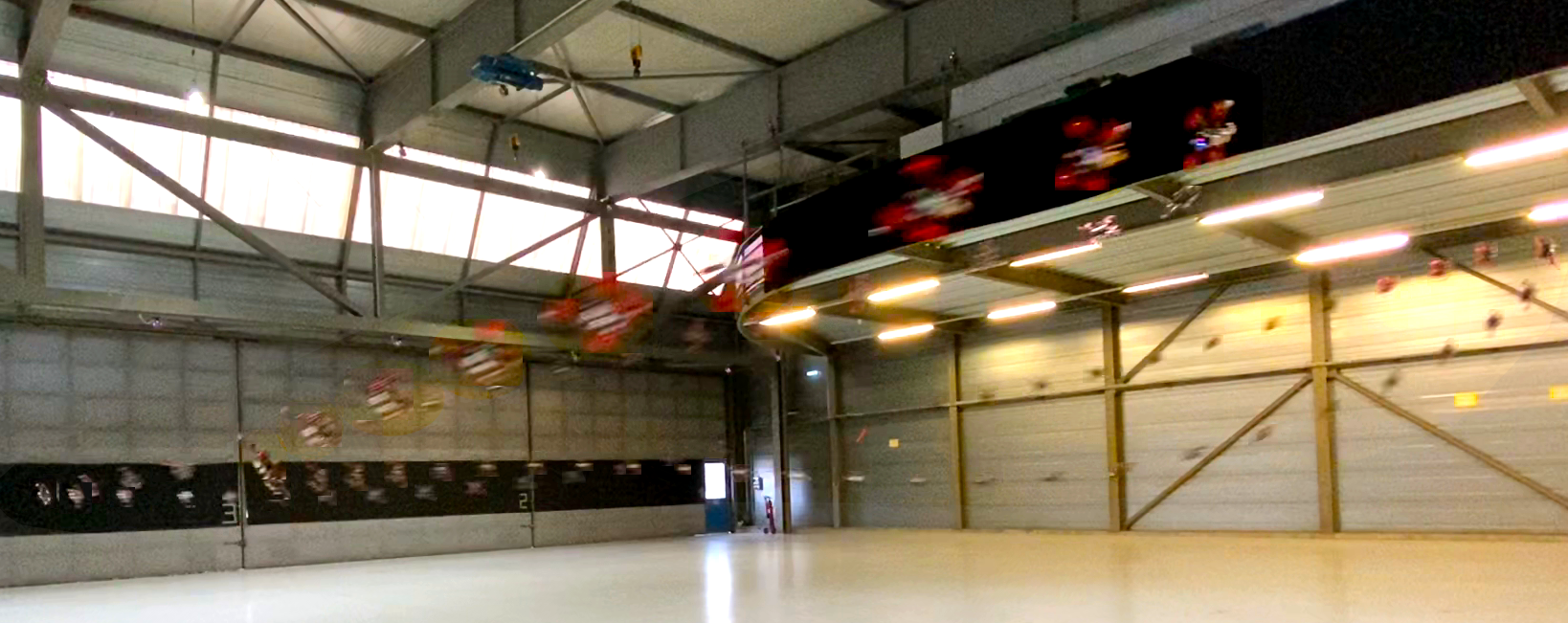}
  \caption{A composite image showing the real-world flight.}
 \label{fig:real_world}
\end{figure}
A reference trajectory is generated by performing a deterministic policy rollout for four continuous laps on the \textit{Split-S} track. 
This reference trajectory is then tracked by a model predictive controller.
Fig.~\ref{fig:real_world} shows an image of the real-world flight. 
Our trajectory enables the vehicle to fly at high speeds of up to~\SI{60}{\kilo\meter\per\hour}, pushing the platform to its physical limits. 
However, executing the trajectory results in large tracking errors. 
Future research will focus on improving the time-optimal trajectory tracking performance. 

%

\section{Conclusion}
In this work, we presented a learning-based method for training a neural network policy that can generate near-time-optimal trajectories through multiple gates for quadrotors. 
We demonstrated the strengths of our approach, including near-time-optimal performance, the capability of handling large track changes, and the scalability and generalizability to tackle large-scale random track layouts while retaining computational efficiency.
We validated the generated trajectory with a physical quadrotor and achieved aggressive flight at speeds of up to~\SI{60}{\kilo\meter\per\hour}.
These findings suggest that deep RL has the potential for generating adaptive time-optimal trajectories for quadrotors and merits further investigation.

\section*{Acknowledgement}
The authors thank Philipp Foehn and Angel Romero for helping with the real-world experiments and providing the trajectory optimization baseline. Also, the authors thank Prof. Jürgen Rossmann and Alexander Atanasyan of the Institute for Man-Machine Interaction, RWTH Aachen University for their helpful advice.

\bibliographystyle{IEEEtran}
\typeout{} 
\bibliography{references}

\end{document}